\documentclass{article}

\usepackage{microtype}
\usepackage{graphicx}
\usepackage{booktabs} % for professional tables
\usepackage{amsmath,amssymb,amsfonts}
\usepackage{algorithm,algorithmic}
\usepackage{graphicx}
\usepackage{textcomp}
\usepackage{xcolor}
\usepackage{mathrsfs}

\usepackage{amsthm}
\usepackage{dsfont}
\usepackage{mathtools}
\usepackage{subfig}
\usepackage{amsmath,amsfonts,amsthm,bm}
\newcommand{\xl}[1]{\textcolor{red}{\red}}
\newcommand{\states}{\mathcal{S}}
\newcommand{\actions}{\mathcal{A}}
\newcommand{\Exp}{\mathds{E}}
\newcommand{\Prob}{\mathds{P}}

\newtheorem{theorem}{Theorem}

\newtheorem{lemma}{Lemma}

\newcommand{\code}[1]{\texttt{#1}}

\DeclareMathOperator*{\argmax}{arg\,max}

\DeclareMathOperator*{\regret}{Regret}

\DeclarePairedDelimiter{\norm}{\lVert}{\rVert}

\usepackage{hyperref}

\usepackage[accepted]{icml2020}
\begin{document}

\twocolumn[
\icmltitle{Opportunistic Episodic Reinforcement Learning}

\icmlsetsymbol{equal}{*}

\begin{icmlauthorlist}
\icmlauthor{Xiaoxiao Wang}{to}
\icmlauthor{Nader Bouacida}{to}
\icmlauthor{Xueying Guo}{to}
\icmlauthor{Xin Liu}{to}
\end{icmlauthorlist}

\icmlaffiliation{to}{University of California, Davis, CA}
\vskip 0.3in
]

\printAffiliationsAndNotice{} 
% \icmlsetsymbol{equal}{*}

\begin{abstract}
In this paper, we propose and study opportunistic reinforcement learning - a new variant of reinforcement learning problems where the regret of selecting a suboptimal action varies under an external environmental condition known as the variation factor. When the variation factor is low, so is the regret of selecting a suboptimal action and vice versa. Our intuition is to  exploit more when the variation factor is high, and explore more when the variation factor is low.  We demonstrate the benefit of this novel framework for finite-horizon episodic MDPs by designing and evaluating OppUCRL2 and OppPSRL algorithms. Our algorithms dynamically balance the exploration-exploitation trade-off for reinforcement learning by introducing variation factor-dependent optimism to guide exploration. We establish an $\tilde{O}(HS \sqrt{AT})$ regret bound for the OppUCRL2 algorithm and  show through simulations that both OppUCRL2  and OppPSRL algorithm outperform their original corresponding algorithms. 
\end{abstract}

\section{Introduction} \label{intro}
Recently, reinforcement learning (RL) has shown spectacular success. By experimenting, computers can learn how to autonomously perform tasks that no programmer could teach them. However, the performance of these approaches  significantly depends on the application domains. In general, we need reinforcement learning algorithms with both good empirical performance and strong theoretical guarantees. This goal cannot be achieved without the efficient exploration of the environment, which has been studied in episodic RL.

In  episodic RL, an agent interacts with the environment in a series of episodes while tries to maximize total reward accumulated over time~\cite{burnetas1997optimal,sutton1998introduction}. This learning process leads to a fundamental trade-off: Shall the agent explore insufficiently-understood states and actions to gain new knowledge resulting in better long-term performance, or exploit its existing information to maximize short-run rewards? The existing algorithms focus on  how  to  balance  such  a  trade-off  appropriately under the implicit assumption that the exploration cost remains  the same  over  time.
However, in a variety of application scenarios, the exploration cost is time-varying  and situation-dependent. Such scenarios can provide us  an opportunity to explore more when the exploration cost is relatively low and exploit more when that cost is high, thus adaptively balancing the exploration-exploitation trade-off to achieve better performance. 
Consider the following motivating examples.

\paragraph{Motivating scenario 1: return variation in game.} 
In a game or a gambling machine where, in some rounds, players may attain special multipliers ($2 \times$, $4 \times$, ..., {etc) on their reward. They can win a large number of points by getting lucky and having large prizes supplemented by large multipliers. Hence, when the player is given a large multiplier, he would better play the move that he believes is the best. Such a conservative move is less risky, especially that we already know that taking a ``bad" action in this situation will result in a significant loss. On the other hand, in a game round with no multiplier or a small one, playing an experimental action will be less risky, since the regret of trying a suboptimal move, in this case, will be lower. 

\paragraph{Motivating scenario 2: value variation in sequential recommendations.} 
For sequential recommender system in e-commerce, the system successively suggests candidate products for users to maximize the total click-through rate (i.e., the probability that a user accepts the recommendation) based on users' preferences and browser history. We note that the real monetary return of a recommendation (if accepted) can differ depending on other factors, such as users with different levels of purchasing power or loyalty (e.g., diamond vs. silver status). Because the ultimate goal is to maximize the overall monetary reward, intuitively, when the monetary return of a recommendation (if accepted) is low, the monetary regret of suggesting suboptimal products is low, leading to a low exploration cost, and correspondingly, high returns lead to high regret and high exploration cost. 

\paragraph{Opportunistic reinforcement learning.}
Motivated by these examples,  we propose and study \emph{opportunistic episodic reinforcement learning}, a new paradigm of reinforcement learning problems where the regret of executing a suboptimal action depends on a varying cost referred to as \emph{variation factor}, associated with the environmental conditions. When the variation factor is low, so is the cost/regret of picking a suboptimal action and vice versa. Therefore, intuitively, we should explore more when the variation factor is low and exploit more when the variation factor is high. As its name suggests, in  opportunistic RL, we leverage the opportunities of variation factor's dynamics to reduce regret. 
\paragraph{Contributions.}
In this work, we propose OppUCRL2 algorithm for opportunistic learning in episodic RL that introduces variation factor-awareness to balance the exploration-exploitation trade-off. The OppUCRL2 can significantly outperforms the UCRL2 \cite{jaksch2010near} in the simulation and have same theoretical guarantee with respect to the regret. The opportunistic RL concept is also easy to generalize for other reinforcement algorithms. To demonstrate it, we design  OppPSRL algorithm based on PSRL  \cite{ian2013more}. It also achieves better performance compared with the original version in the simulation. To the best of our knowledge, this is the first work proposing and studying the concept of the opportunistic reinforcement learning. 
% Our algorithms Opportunistic UCRL2 and  OppPSRL significantly outperforms than the corresponding  original version.
We believe this work will serve as a foundation for the opportunistic reinforcement learning concept and help further addressing the exploration-exploitation trade-off. 

\section{Related Work}

Optimism in the face of uncertainty (OFU) is a popular paradigm for the exploration-exploitation trade-off in RL, where each pair of states and actions is offered some optimism bonus. The agent then chooses a policy that is optimal under the ``optimistic" model of the environment. To learn efficiently, it maintains some control over its uncertainty by assigning a more substantial optimistic bonus to potentially informative states and actions. The assigned bonus can stimulate and guide the exploration process. Most OFU algorithms provided strong theoretical guarantees~\cite{azar2017minimax,bartlett2009regal,jaksch2010near,dann2015sample,strehl2009reinforcement}. A popular competitor to OFU algorithms is inspired by Thompson sampling (TS)~\cite{chapelle2011an}. In RL, TS approaches~\cite{strens2000bayesian} maintain a posterior distribution over the reward function and the transition kernel, then compute the optimal policy for a random sampled MDP from the posterior. 
One of the well-known TS algorithms in the literature is Posterior Sampling for Reinforcement Learning (PSRL)~\cite{ian2013more,ian2017why}. 

The opportunistic learning idea has been introduced in \cite{Wu2018_AdaUCB} for classic $K$-armed bandits and in \cite{Guo2019AdaLinUCBOL} for context bandits. In the reinforcement learning, the authors in  \cite{dann2019policy} consider the case where the each episode has a side context and propose ORLC algorithm that can use the context information to estimate the dynamic of the environment but not include the opportunistic concept, which is different from us.
To the best of our knowledge,
no prior work has made formal mathematical formulation and rigorous performance analysis for opportunistic reinforcement learning.

\section{Problem Formulation} \label{probForm}

We consider an RL problem in an episodic finite-horizon Markov decision process (MDP), $M:=\langle \mathcal{S},\mathcal{A},H, P, r \rangle$, where $\mathcal{S}$ is a finite state space with carnality  $|\mathcal{S}|=S$,  $\mathcal{A}$ is a finite action space with carnality $|\mathcal{A}|=A$,  $H$ is the horizon that represents the number of time steps in each episode, $P$ is a state transition distribution such that $P( \cdot |s,a)$ dictates a distribution over state $\mathcal{S}$ if action $a$ is taken for state $s$, and $r: \mathcal{S}\times \mathcal{A}\rightarrow [0,1]$ is the deterministic reward function. 
For simplicity, we assume that
the reward function $r$ is known to the agent but the transition distribution $P$ is unknown.

 In each episode of this MDP, an initial state $s_1 \in \mathcal{S}$ is chosen arbitrarily by the environment before it starts. For each step $h \in [H]$\footnote{We write $[n]$ for ${i\in \mathbb{N},1\leq i \leq n}$}, the agent observes a state $s_h \in \mathcal{S}$,  selects an action $a_h \in \mathcal{A}$, receives a reward $r(s_h,a_h)$ and then the state transits to next state $s_{h+1} \in \mathcal{S}$ that is drawn from the distribution $P( \cdot |s_h,a_h)$. The episode ends in state $s_{H+1}$.

A policy for an agent during the episode is expressed as a mapping $\pi: \states \times [H] \rightarrow \actions$.  We write $V^{\pi}_{h}: \states \rightarrow \mathbb{R}$ as the value function at step $h$ under policy $\pi$. For a state $s \in \mathcal{S}$, $V^{\pi}_{h}(s)$ is the expected return (i.e., sum of rewards) received under policy $\pi$, starting from $s = s_h \in \states$, i.e., $V^{\pi}_{h}(s) :=  \Exp\left[ \sum_{i=h}^{H} r(s_i, \pi(s_i,i)) \Big| s_h = s \right ]$. Because the action space, state space, and horizon are finite, and the reward function is deterministic, there always exits an optimal policy $\pi^*$ that attains the best value $V_h^*(s) = sup_{\pi} V_h^\pi (s)$ for all $s\in \mathcal{S}$ and $h \in [H]$. For an episode with initial state $s_1$, the quality of a policy $\pi$ is measured by the regret that is the gap between the value function at step $1$ under policy $\pi$ and that under optimal policy, i.e., $V^*_1(s_{1}) - V^{\pi}_1(s_{1})$. The goal of the classic RL problem is to consider a RL agent interacts with the environment (MDP $M$) for $K$ episodes $k\in [K]$ in a sequential manner and find the optimal policy.

Next we introduce  the \textbf{opportunistic reinforcement learning} in an episodic finite-horizon MDP. For each episode $k \in [K]$, let $L_k \geq 0$ be an external \textbf{variation factor} and  not change during the episode. We assume $L_k$  is independent of the MDP $M$ for $k \in [K]$. To distinguish different episodes, we use $s_{k,h}$, $a_{k,h}$, $r_{k,h}$ to denote the state, action and reward in step $h$ of episode $k$. The \textbf{expected actual return} for the episode $k$ is defined as $\mathds{E}[L_k V_1^\pi(s_{k,1})]$ if the initial state is {$s_{k,1}$} and the policy of agent is $\pi$.  Before the $k$-th episode, the agent can observe the initial state $s_{k,1}$ and the current value of $L_k$. Based on the policy $\pi_k$ that the agent selected, the expected actual return that the learner receives is $\mathds{E}[L_k V_1^\pi(s_{k,1}])$.

This model captures the essence of the opportunistic RL paradigm for the motivating scenarios in introduction. In the opportunistic RL model, we notice that the optimal policy that maximize $\mathds{E}[L_k V^{\pi_k}_h(s_{k,1})]$ for each episode $k \in [K]$ does not change over episodes and is same as the optimal policy $\pi*$  in the standard RL problem for a MDP $M$. So, the best expected actual return for an episode $k$ is $\mathds{E}[L_k V^{*}_h(s_{k,1})]$. 
 
 The goal is to minimize the \textbf{actual total regret} for $K$ episodes in terms of the  expected actual return. Particularly, we define the actual total regret in opportunistic RL problem over  $K$ episodes regarding the expected actual return as: 

\begin{equation} \label{regret}
\setlength{\abovedisplayskip}{0pt} %%% 
\setlength{\belowdisplayskip}{-1pt}
Regret(K) := \sum_{k=1}^{K}\mathds{E} \left[  L_kV^*_1(s_{k,1}) -L_k V^{\pi_k}_1(s_{k,1}) \right]
\end{equation}

In a special case, equation \eqref{regret} has an equivalent form: when $L_k$ is i.i.d. over the episodes with mean value $\bar{L}$, the total regret regarding actual reward is $\regret(K) = \bar{L} \sum_{k=1}^{K} V^*_1(s_{k,1}) -\sum_{k=1}^{K}\mathds{E}[L_k V^{\pi_k}_1(s_{k,1})]$.  
Note that in general, it is likely that $ \mathbb{E}[L_k V^{\pi_k}_1(s_{k,1})]\neq \bar{L} \mathbb{E}[  V^{\pi_k}_1(s_{k,1})]$, because the policy $\pi_k$ can depend on $L_k$.

\section{Opportunistic Reinforcement Learning  Algorithm}

In this section, we propose two opportunistic algorithms that are designed based on the optimism in the face of uncertainty and the posterior sampling respectively.

We first introduce OppUCRL2 algorithm, an opportunistic variant of UCRL2~\cite{jaksch2010near}.

In Alg.\ref{alg:ucrl2},  $\delta \in(0,1]$ is a hyper-parameter, and $\tilde{L}_k$ is the normalized variation factor, defined as,

 \begin{equation} \label{loadexp}
 \setlength{\abovedisplayskip}{3pt} %%% 
\setlength{\belowdisplayskip}{3pt}
\tilde{L}_k = \frac{[L_k]_{l_{\min}}^{l_{\max}} - l_{\min}}{l_{\max} -l_{\min}}
\end{equation}
where $l_{\min}$ and $l_{\max}$ are respectively the lower and upper thresholds for truncating the variation factor level, and $[L_k]_{l_{\min}}^{l_{\max}} = \max\{l_{\min}, \min\{L_k,  l_{\max}\} \}$. The variation factor normalization restricts the impact of the variation factor term in the confidence bounds, which avoids under or over explorations. We note that the normalized variation factor $\tilde{L}_k$ is only employed in the algorithm itself. Indeed, the regret depends on the real variation factor $L_k$ and not $\tilde{L}_k$.

In the initialization, $N(s,a)$ and  $N(s,a,s')$  are  the counts for state-action pair $(s,a)$ played and tuple $(s,a,s)$ happened up to current episode. $t_k= H(k-1)$ is the start time of the episode $k$.  Before the start of the $k$-th episode, the algorithm  observes the variation factor $L_k$ and normalize it by Eq.\ref{loadexp} in Line 4. The empirical estimate $\hat{P}_k(\cdot|s,a)$ of $P(\cdot|s,a)$ is calculated by all historical transitions observed so far in Line 6. The width of the high probability confidence regions of $\hat{P}(\cdot|s,a)$ is estimated by Hoeffding's inequality and normalized variation factor $\tilde{L}_k$ in line 7. Then,  a plausible MDP set $\mathcal{M}_k$ is created in line 8 that consists of finite-horizon MDP $M'$ with same known reward function $r$ and the transition probability $P'(\cdot|s,a)$  in the high probability confidence regions of $\hat{P}(\cdot|s,a)$ with width $d_k(s,a)$ for all state-action pairs. 

\setlength{\textfloatsep}{0.1cm}
\setlength{\floatsep}{0.1cm}
\begin{algorithm}

   \caption{OppUCRL2}
   \label{alg:ucrl2}
\begin{algorithmic}[1]
\STATE{\textbf{Input}: $l_{min},l_{max}, \delta$}
   \STATE {\textbf{Initialization}: $ N(s,a,s')=0$, $N(s,a)=0$  $\forall, s\in \mathcal{S}, a \in \mathcal{A}, s' \in \mathcal{S}$}
   \FOR{\text{episode} $k=1,2,\dots, K$}
        \STATE Observe $L_k$ and calculate $\tilde{L}_k$ by Eq. \ref{loadexp} \label{line:l}
        \STATE $ t_k = H(k-1)$
        % \STATE {\textbf{Empirical estimation for transmit probability }}
        \STATE $\hat{P}_k(s'|s,a) = \frac{N(s,a,s')}{N(s,a)}$ \label{line:p}
        \STATE $d_k(s,a)  = \sqrt{\frac{2 S (1-\tilde{L}_k) \log(2 S A t_k / \delta)}{\max\left \{ 1, N(s,a) \right \}}} \label{bound2}$\label{line:w_p}
        \STATE { \label{line:set}
        $\mathcal{M}_k := \{ M':  \norm[\big]{ P'(.|s,a) - \hat{P}_k(.|s,a) }_1 \leq d_k(s,a)$ $
		\forall (s,a)\in \mathcal{S}\times\mathcal{A} \}$
        }
        % \STATE {\textbf{Optimistic planning}}
%        \STATE ${\pi}_k, \tilde{M_k} \leftarrow Alg.2(\hat{P}_k,d_k)$ \label{line:evi}
        
          \STATE ${\pi}_k, \tilde{M_k} \leftarrow ExtendedValueIteration(\mathcal{M}_k)$ \label{line:evi}
       
        % \STATE {\textbf{Execute policy $\pi_k$}}
        
        \FOR {\text{time step} $h = 1,\ldots,H$} \label{line:ex_start}
            \STATE $a_{k,h} = \pi_k(s_{k,h})$.
            \STATE Observe $r_{k,h}$ and $s_{k,h+1} \sim P(\cdot|s_{k,h},a_{k,h})$.
%            \STATE {\textbf{Update empirical estimation}}
            \STATE $N(s_{k,h},a_{k,h},s_{k,h+1})\! \leftarrow \! N(s_{k,h},a_{k,h},s_{k,h+1})\! +\! 1$
            \STATE $N(s_{k,h},a_{k,h}) \leftarrow N(s_{k,h},a_{k,h}) + 1$
        
       \ENDFOR
   \ENDFOR
\end{algorithmic}
\end{algorithm}

Next, in line 9, Alg.\ref{alg:ucrl2} calls a subroutine Finite Horizon Extended Value Iteration (see Appendix~\ref{se:ap_extend_value_iteration}  for more details) that returns an optimistic MDP $\tilde{M}_k$ with the best achievable reward from $\mathcal{M}_k$ and the optimistic policy $\pi_k$. The idea behind finite horizon extended value iteration is same as ~\cite{puterman1994mdp,dann2015sample}.
Last, the policy $\pi_k$ executes throughout the episode $k$ adn updates the counts $N(s,a,s')$ and $N(s,a)$.

In general,  Alg.\ref{alg:ucrl2} explores more when the variation factor is relatively low, and exploits more when the variation factor is relatively high. To see this, note that $d_k(s,a)$ in line 7 is the adaptive width of the confidence region modulated by $\tilde{L}_k$ for MDP set $\mathcal{M}_k$, which determines the level of exploration. For example, when $L_k$ is at its lowest level with $L_k \leq l_{min}$,  $\tilde{L}_t=0$, and the width of confident region $d^k$ is the same as that of the UCRL2 algorithm, and then the  algorithm learns the policy in the same way as the conventional UCRL2. At the other extreme, when $\tilde{L}_k = 1$, i.e., $L_k \geq l_{max}$, the width $d_k=0$, that is, when the variation factor is at its highest level, the  algorithm purely exploits the existing knowledge and selects the best policy. With the exploitation of variation factor awareness capabilities and given that the actual regret is scaled with the variation factor level, OppUCRL2 could achieve a lower regret than the original UCRL2. 

Similarly, we also can generalize the opportunistic RL concept into the sampling based algorithm,  OppPSRL, which is a variant of Posterior Sampling for Reinforcement Learning (PSRL)~\cite{ian2017why}. (See Appendix~\ref{se:ap_OppPSRL}  for more details).

\section{Regret Analysis for OppUCRL2}

\begin{figure*}[!h]

\captionsetup{belowskip=0pt,captionskip=01pt}
	\centering
	\subfloat[\emph{River Swim} \label{rs1}]{\includegraphics[height = 0.23\textwidth,width=0.333\textwidth]{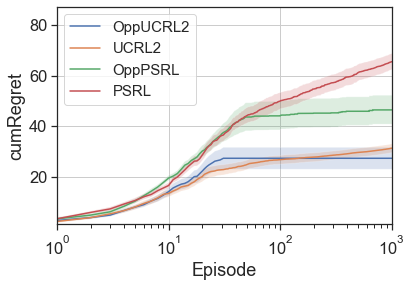}}\hfill
	\subfloat[\emph{Cliff Walking} \label{cw1}]{\includegraphics[height = 0.23\textwidth,width=0.333\textwidth]{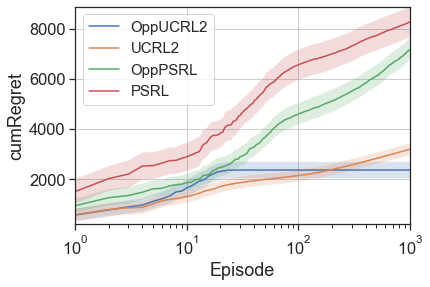}}\hfill
	\subfloat[\emph{Frozen Lake} \label{fl1}]{\includegraphics[height = 0.23\textwidth,width=0.333\textwidth]{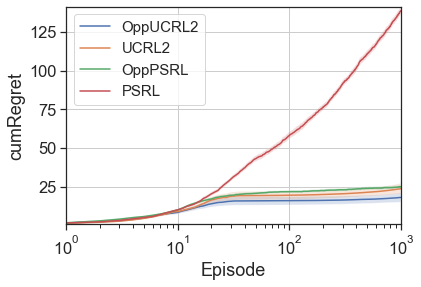}}\hfill

	\caption{Regret under binary variation factor scenarios }
	\label{BinaryRegret}
\end{figure*}

In this section, we present an upper bound on the regret of OppUCRL2. We study a simple case with periodic square wave variation factor. Specifically, we assume that the variation factor for an episode $k$ is $L_k=\epsilon_0$ if the episode index $k$ is even, and $L_k=\epsilon_1$ if $k$ is odd.
 Because we use a sophisticated variation factor-aware regret expression as described in Eq.~\ref{regret} for the opportunistic learning that is different from the classical regret definition, in order to compare OppUCRL2 and original UCRL2 algorithm fairly, we should derive the regret bounds for both of them based on Eq.~\ref{regret}. Following the same logic  as \cite{jaksch2010near,glp2020aaaitutorial}, we can get Theorem \ref{the: pswlucrl2} and Theorem \ref{the: opucrl2} that show UCRL2 and OppUCRL2  can achieve the same regret bound $\tilde{O}(HS\sqrt{AT})$ in the periodic square wave variation factor case.
(see Appendix~\ref{se:ap_analysis}  for more details).

\begin{theorem}[Regret Bound for UCRL2 under Periodic Square Wave Variation Factor]
\label{the: pswlucrl2}
For a finite  horizon MDP, $M := \langle \mathcal{S},\mathcal{A},H, P, r \rangle$, and $L_k = \epsilon_0$ if the episode index $k$ is even, and $L_k = 1- \epsilon_1$ if $k$ is odd, consider a parameter $\delta$, then the regret of UCRL2 is bounded with a probability at least $1-\delta$ by,
\begin{equation*}
\setlength{\abovedisplayskip}{2pt} %%% 
\setlength{\belowdisplayskip}{2pt}
Regret(K) = \tilde{O}(HS\sqrt{CAT})
\end{equation*}
where $C = \log(2 S A T / \delta)$.
\end{theorem}

\begin{theorem}[Regret Bound for OppUCRL2 under Periodic Square Wave Variation Factor]
\label{the: opucrl2}
For a finite  horizon MDP, $M := \langle \mathcal{S},\mathcal{A},H, P, r \rangle$, and $L_k = \epsilon_0$ if the episode index $k$ is even, and $L_k = 1- \epsilon_1$ if $k$ is odd, consider a parameter $\delta$, then the regret of OppUCRL2 is bounded with a probability at least $1-\delta$ by, 
\begin{equation*}
\setlength{\abovedisplayskip}{2pt} %%% 
\setlength{\belowdisplayskip}{2pt}
Regret(K) = \tilde{O}(HS\sqrt{CAT})
\end{equation*}

where $C = \log(2 S A T / \delta)$.
\end{theorem}

\section{Experimental Evaluation}

In this section, we evaluate the empirical performance of OppUCRL2 and OppPSRL compared to the original UCRL2 and PSRL algorithms. 
We use three classic examples of the OpenAI Gym, namely River Swim, Cliff Walking and Frozen Lake that represent three different test cases~\cite{strehl2008analysis}: undiscounted reward in a stochastic environment,  undiscounted reward deterministic environment, and discounted reward in deterministic environment (see Appendix~\ref{se:ap_environment}  for more details). The stochastic and deterministic describes the state transition distribution. The River Swim and Cliff Walking RL environments can be formulated as an undiscounted, episodic MDPs while Frozen Lake is a discounted, episodic task with a discount factor $\gamma  = 0.95$.  We report the results for the average of 20 simulations with different seeds while showing  95\% confidence interval.  We use the same scaling factors for both algorithms, chosen experimentally for each environment.
We compare all algorithms with its best input precision hyper-parameters obtained by grid search.

We first introduce the result under random binary-valued variation factor. We assume that the variation factor $L_k$ is i.i.d. over the episodes, with $L_k \in\{\epsilon_0, 1- \epsilon_1 \}$, where $\epsilon_0, \epsilon_1\geq 0$ and $\epsilon_0<1-\epsilon_1$. Let $\rho$ denote the probability that the  variation factor is low, i.e., $\mathbb{P}\{L_k = \epsilon_0 \}=\rho$. Fig. \ref{BinaryRegret}  shows  the  regret  for  different  algorithms  under random binary-value variation factor with $\epsilon_0 = \epsilon_1 = 0$ and $\rho = 0.5$.

 Opportunistic RL algorithms outperform the corresponding original RL algorithms across every environment by significantly reducing the regret.  More significantly, for River Swim, Cliff Walking and Frozen Lake, at the end of the $10^3$-th episode, OppUCRL2 reduces the regret by $12.7\%$, $25.9\%$ and $23.7\%$  respectively compared with UCRL2. OppPSRL reduces the regret by $29.1\%$, $13.3\%$ and $81.9\%$ respectively compared with PSRL. 
 We also notice that OppUCRL2 achieves $O(1)$ regret converging to the optimal policy in a constant time. This is because it pushes most exploration moves to the episodes where the variation factor is equal to zero. As a result, the exploration cost is negligible. Although OppUCRL2 largely outperforms UCRL2, it may have higher regret at the beginning, especially in the environments with less deterministic behavior such as River Swim and Cliff Walking. OppUCRL2 emphasizes the main insight of exploration-exploitation trade-off: we may sacrifice some short-term rewards to improve future performance. This observation combined with the constant-time optimal regret demonstrates OppUCRL2 capability to learn and adapt to the environment's  dynamics overtime. We also test these algorithms in the continuous variation factor case and find the opportunistic version of the algorithms also have a better performance (see Appendix~\ref{se:ap_beta}  and \ref{se:ap_discussion} for more details).

\section{Conclusion}
In this paper, we study opportunistic reinforcement learning where the regret of choosing a suboptimal action depends on an external condition denoted variation factor. We establish OppUCRL2 and OppPSRL algorithms,  variants for the well-known UCRL2 and PSRL algorithms. We also analyze the regret of OppUCRL2 and  present $\tilde{O}(HS \sqrt{AT})$ regret bounds.  Experimental results demonstrate substantial benefits from employing low-cost opportunistic exploration in OppUCRL2 and OppPSRL algorithm  under variation factor fluctuations. 

\bibliography{example_paper}
\bibliographystyle{icml2020}

\newpage
\appendix

\section{Algorithms}\label{se:apalg}
\subsection{Finite Horizon Extended Value Iteration}
\label{se:ap_extend_value_iteration}
\begin{algorithm}[!h]
   \caption{Finite Horizon Extended Value Iteration}
   \label{alg:exten1}
   
\begin{algorithmic}[1]
    \STATE \textbf{Input}: MDP set $\mathcal{M}$

    \STATE \textbf{Initialize} \scalebox{0.9}{$V_{H+1}(s) = 0$} for all  \scalebox{0.9}{$s \in \states$}
    
    \FOR{\scalebox{0.9}{$h=H,H-1,\dots,1$}}
            \STATE  Sort the states in $\states$ in the descending order w.r.t. their values: Let \scalebox{0.9}{$\states = \{s'_1, s'_2, ..., s'_S\}$} such that \scalebox{0.9}{$V_{h+1}(s'_1) \geq V_{h+1}(s'_2) \geq \dots \geq V_{h+1}(s'_S)$}
        \FOR {\scalebox{0.9}{$(s,a) \in \states \times \actions$}}

            \STATE \scalebox{0.9}{$\tilde{P}(s'_1|s,a) = \min\left\{1, \hat{P}(s'_1|s,a) + \frac{d(s,a)}{2} \right\}$}
            \STATE \scalebox{0.9}{$\tilde{P}(s'_i|s,a) = \hat{P}(s'_i|s,a) $ for all $1 < i \leq S$}
            \STATE \scalebox{0.9}{Set $j = S$}
            \WHILE {\scalebox{0.9}{$\sum_{s'_i \in \states}  \tilde{P}_k(s'_i|s,a) > 1$}}
                \STATE \scalebox{0.9}{$\tilde{P}(s'_j|s,a) = \max\{0, 1-\sum_{s'_i \neq s'_j}  \tilde{P}(s'_i|s,a)\}$}
                \STATE \scalebox{0.9}{$j = j-1$}
            \ENDWHILE
            \STATE \scalebox{0.9}{$Q_{h}(s,a) = r(s,a) + \sum\limits_{s' \in \states} \tilde{P}(s'|s,a) V_{h+1}(s')$}

        \ENDFOR
             \STATE \scalebox{0.9}{$V_{h}(s) = \max_{a \in \actions} Q_{h}(s,a)$}
    \ENDFOR
    
    \STATE $\pi(s,h) = \arg \max_{a\in \mathcal{A}}Q_h(s,a) $ for  all $s \in \mathcal{S}$, $h \in [H]$
    
    \STATE {\bfseries Output:} MDP with transition probabilities $\tilde{P}$, and  optimal policy $\pi$
\end{algorithmic}
\end{algorithm}

Alg.\ref{alg:exten1}  Finite Horizon Extended Value Iteration is used as a subroutine for Alg.\ref{alg:ucrl2} OppUCRL2. The input of Alg.\ref{alg:ucrl2} is a MDP set. The output is 
the optimistic MDP $\tilde{M}_k$ with the best achievable reward from $\mathcal{M}_k$ and the optimistic policy $\pi_k$ in line \ref{line:evi}. In practical , we define the value function in finite horizon MDP $M'$ under policy $\pi$ as $V^{M'(\pi)}_{h}(s) :=  \Exp_{p(\cdot|s,a) \sim M'}[ \sum_{i=h}^{H} r(s_i, \pi(s_i,i))| s_h = s ]$. Then the optimistic MDP $\tilde{M}_k$ and optimistic policy $\pi_k$ are $\tilde{M}_k,\pi_k =\arg \max_{M' \in \mathcal{M}_k,\pi} V^{M'(\pi)}_{1}(s)$ for all $s \in \mathcal{S}$. The idea behind finite horizon extended value iteration is same as ~\cite{puterman1994mdp,dann2015sample}, putting as much transition probability as possible to the state with maximal value at the expense of transition probabilities to states with small values. Then, in order to make $\tilde{P}$ correspond to a probability distribution again, the transition probabilities with small values are reproduced iteratively with respect to the constraint $d(s,a)$. This implies that extended value iteration solves a linear optimization problem over the convex polytope constructed by the set of transition probabilities satisfying conditions and $d(s,a)$.

\subsection{OppPSRL}
\label{se:ap_OppPSRL}
In this section, we generalize the opportunistic RL concept into the sampling based algorithm,  OppPSRL, which is a variant of Posterior Sampling for Reinforcement Learning (PSRL)~\cite{ian2017why}. In each episode, PSRL samples a single MDP from the plausible MDP set and 
then selects a policy that has maximum value for that MDP.
\begin{algorithm}
   \caption{\code{OppPSRL}}
   \label{alg:psrl2}
\begin{algorithmic}[1]
\STATE{\textbf{Input}: prior distribution   $\phi(\bm{\alpha}_{0,1})$ of $M$} 
   \FOR{\text{episode} $k=1,2,\dots, K$}
        \STATE Observe $L_k$ and calculate $\tilde{L}_k$ by Eq.\ref{loadexp}
        \STATE $\bm{\alpha}_k =  \tilde{L}_k\bm{\alpha}_{k-1,1}$
        \STATE Sample MDP $M_k \sim  \phi(\cdot |\bm{\alpha}_k)$
        \STATE Compute $\pi_k = \argmax_\pi V_{1}^{M_k{(\pi)}}$
        \FOR {\text{time step} $h = 1,\ldots,H$} \label{line:ex_start}
            \STATE $a_{k,h} = \pi_k(s_{k,h})$.
            \STATE Observe $r_{k,h}$ and $s_{k,h+1} \sim P(\cdot|s_{k,h},a_{k,h})$.
            \STATE Update the parameters $\bm{\alpha}_{k,h}$ of posterior distribution by $ (s_{k,h},a_{k,h},r_{k,h}, s_{k,h+1})$.
       \ENDFOR
   \ENDFOR
\end{algorithmic}
\end{algorithm}

Inspired by the opportunistic learing idea, we propose  Alg.\ref{alg:psrl2} OppPSRL. The input is a prior distribution of the MDP. In general, we can formulate the state transition distribution $P( \cdot |s,a)$ as a Dirichlet distribution $\phi$ with parameters $\bm{\alpha}$. At the begin of each episode, 
 Alg. \ref{alg:psrl2} calculates the normalized variation factor $\tilde{L}_t$ in line 3. Then it uses  $\tilde{L}_k$ to rescale the parameter $\bm{\alpha_k}$. Next, a MDP $M_k$ is sampled from the scaled distribution. In line 6, it  computes the policy $\pi_k$ that has maximum value for the MDP. Finally, the policy $\pi_k$ is executed throughout the episode $k$ and the posterior distribution is updated by the new observations.  
 
 The core step of  Alg. \ref{alg:psrl2} is Line 4.  Intuitively, when $\tilde{L}_k$ is small, it can decrease the value of $\bm{\alpha_k}$ and the corresponding Dirichlet distribution is more concentrated, then the sampled MDP in Line 5 is similar to the empirical MDP with high probability, so the policy $\pi_k$ in Line 6 is more conservative and less exploratory.
 When  $\tilde{L}_k$ is larger, the distribution flattens, it provides the opportunity for the agent to explore new MDP and try under-explored actions.
\section{Regret Analysis}
\label{se:ap_analysis}
This section introduce the proof of the theorems in the main paper.

According to Theorem 2 in \cite{jaksch2010near} and Theorem 1 in \cite{glp2020aaaitutorial}, we have the following lemma that shows the regret bound for UCRL2 in finite horizon MDP.

\begin{lemma}
[Regret Bound for  UCRL2 in finite  horizon MDP]
\label{the: pswlucrl2}
For a finite  horizon MDP, $M := \langle \mathcal{S},\mathcal{A},H, P, r \rangle$, 
consider a parameter $\delta$, the regret of UCRL2 is bounded with a probability at least $1-\delta$ by, 
\begin{align*}
Regret(K) = \tilde{O}(HS\sqrt{CAT})
\end{align*}
where $C = \log(2 S A T / \delta)$.
\end{lemma}
\subsection{Proof of Theorem 1}

\begin{proof}
For the periodic square wave variation factor case, we can categorize the episodes into two groups, then analyze the regret independently, which can still guarantee an upper bound for the regret because the variation factor is independent from the MDP and UCRL2 algorithm. Specifically, from Eq. \ref{regret}, we have 
\begin{align}
\label{eq:deco}
Regret(K) &= \sum_{k\in [K]}\mathds{E} \left[  L_kV^*_1(s_{k,1}) -L_k V^{\pi_k}_1(s_{k,1}) \right ] \nonumber\\ 
&= \sum_{k\in [K], \text{k is odd}}\mathds{E} \left[  L_kV^*_1(s_{k,1}) -L_k V^{\pi_k}_1(s_{k,1}) \right] +\nonumber\\
& \sum_{k\in [K], \text{k is even}}\mathds{E} \left[  L_kV^*_1(s_{k,1}) -L_k V^{\pi_k}_1(s_{k,1}) \right ] \nonumber\\
&= \epsilon_0 * Regret_{original, odd}(K) + \nonumber\\
& (1-\epsilon_1)*Regret_{original, even}(K)
\end{align}
 So, based on the union bound and Lemma 1, we can get the bound  in  Theorem 1.
\end{proof}

\subsection{Proof of Theorem 2}
\begin{proof}
In order to bound the regret of OppUCRL2, we can still do the same decomposition as Eq. \ref{eq:deco}. The difference is that the exploration in OppUCRL2 related to the variation factor, so we cannot directly apply lemma 1 for analysis. However, we notice that in the analysis of UCRL2 in finite horizon MDPs, the regret bound is mainly dominated by the time of visits for all state-action pairs. In order to get the upper bound of the regret of OppUCRL2 in periodic square wave variation factor, we can regard it as two independent algorithms with different exploration parameters in odd and even episodes to get the upper bound of the regret, because the time of visits for each state-action pairs in OppUCRL2 is at least same as that in two independent UCRL2 cases and the difference only affects the constant coefficient in the bound. So, OppUCRL2 can achieve the bound shown in Theorem \ref{the: opucrl2}.
\end{proof}

Theorem \ref{the: pswlucrl2} and Theorem \ref{the: opucrl2} show that UCRL2 and OppUCRL2 in the periodic square wave variation factor can achieve the same regret bound $\tilde{O}(HS\sqrt{AT})$.

\section{Simulation}

\subsection{Environment Setting}
\label{se:ap_environment}

\begin{figure}[H]
	\centering
	\includegraphics[width= 0.8 \columnwidth]{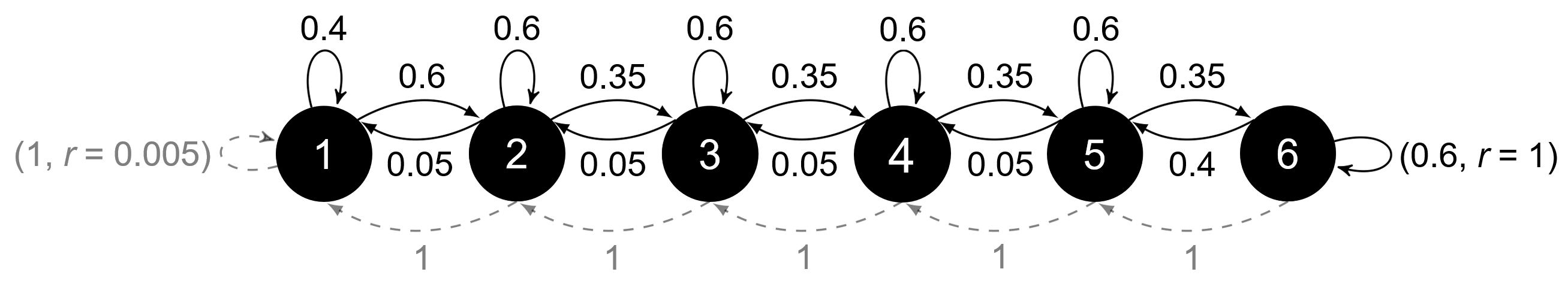}
	\caption{\emph{River Swim} - consisting of six states arranged in a chain and two actions. Continuous and dotted arrows represent the MDP under the actions ``right" and ``left", respectively. The agent always starts in state 1. Swimming left (with the current) is always successful, but swimming right (against the current) often fails. The agent receives a small reward for reaching the starting state, but the optimal policy is to attempt to swim right and receive a much larger reward. We set the horizon $H = 15$ and length of the chain $S = 6$.}
	\label{RiverSwim}
\end{figure}
\begin{figure}[h]
	\centering
	\includegraphics[width= 0.7 \columnwidth]{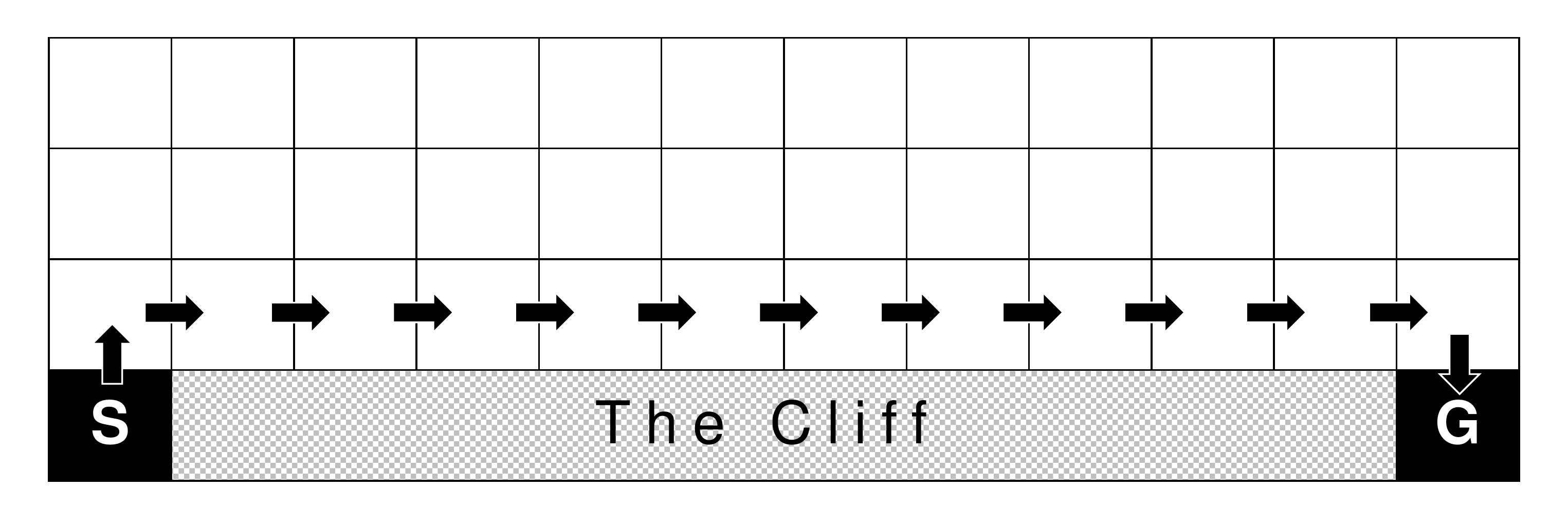}
	\caption{\emph{Cliff Walking} - consisting of a grid. The start state is in the left lower corner while the goal state is in the right lower corner of the grid. The possible actions causing movement are UP, DOWN, RIGHT, and LEFT. Each transition incurs -1 reward, except for stepping into the gray region marked ``The Cliff", which incurs -100 reward and a reset to the start. An episode terminates when the agent reaches the goal or the episode time expires (reach the horizon). A suboptimal policy can be thought of as avoiding moving closely to ``The Cliff" region. The optimal policy is the shortest path avoiding ``The Cliff" region as shown in the figure. We set the horizon $H = 50$ and the shape of grid as $4\times 12$.}
\label{CliffWalking}
\end{figure}
\begin{figure}[h]
	\centering
	\includegraphics[width= 0.3 \columnwidth]{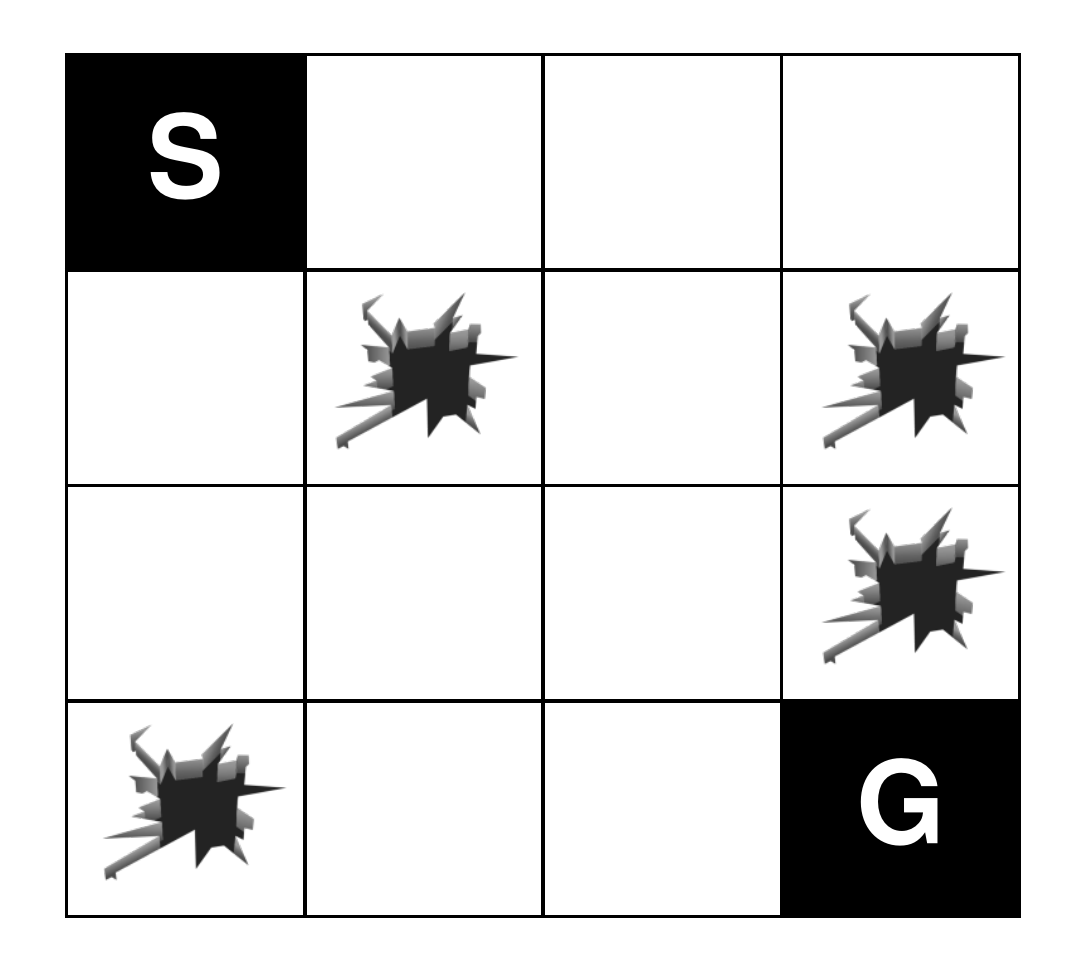}
	\caption{\emph{Frozen Lake} -  consisting of a grid world representing a frozen lake. The water is mostly frozen, but there are a few holes where the ice has melted. Thus, some tiles of the grid are walkable, and others lead to the agent falling into the water. The agent is rewarded for finding a walkable path to a goal tile. The episode ends when it reaches the goal, fall in a hole or the episode time expires. The agent receives a reward of 1 if it reaches the goal, -1 if it falls in a hole and zero otherwise.We set the horizon $H = 20$ and the shape of grid as $4\times 4$.}
	\label{FrozenLake}
\end{figure}
\begin{figure*}[h]
	\centering
	\subfloat[\emph{River Swim} \label{rs2}]{\includegraphics[width=0.333\textwidth]{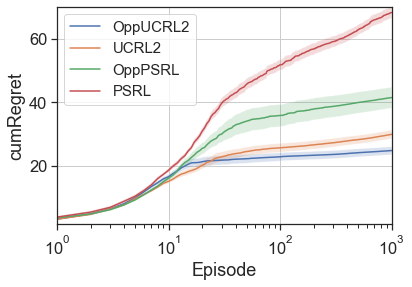}}\hfill
	\subfloat[\emph{Cliff Walking} \label{cw2}]{\includegraphics[width=0.333\textwidth]{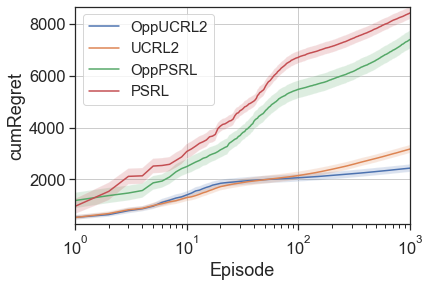}}\hfill
	\subfloat[\emph{Frozen Lake} \label{fl2}]{\includegraphics[width=0.333\textwidth]{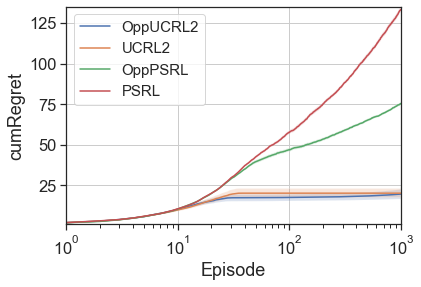}}\hfill
	\caption{Regret under Beta \emph{variation factor} Scenarios }
	\label{BetaRegret}
\end{figure*}
\subsection{Evaluation Using Continuous Variation Factor}
\label{se:ap_beta}
We investigate the performance of the algorithms under continuous variation factor. We assume that the variation factor $L_k$ is i.i.d. over episodes and sampled from a Beta distribution, i.e., $L_k \sim Beta(2,2)$ . 
Figure~\ref{BetaRegret} shows the regrets for different algorithms and environments. Here, we define the lower threshold $l_{min}$ such that $\Prob(L_k \leq l_{min}) = \rho$ where $\rho = 0.05$, and the upper threshold $l_{max}$ such that $\Prob(L_k \geq l_{max}) = \rho$. 

For River Swim, Cliff Walking and Frozen Lake, at the end of the $10^3$-th episode,  OppUCRL2 reduces regret by $17.1\%$, $23.3\%$ and $2.7 \%$ respectively compared wiht UCRL2. OppPSRL reduces regret by $39.2\%$, $12.2\%$ and $43.4\%$ respectively compared with PSRL. 
For OppUCRL2, we also see similar trends with previous experiments. However, with beta variation factor, OppUCRL2 algorithm does not achieve a constant-time regret.  This is due to the fact that the variation factor does not vary radically between 0 and 1, and the exploration carried out in the low variation factor episode usually does not have a zero variation factor, thus, generating an extra overhead compared to the previous experimental case.

\section{Discussion}
\label{se:ap_discussion}
We reserve this section to discuss the limitations of our work and possible future improvements. 

\textbf{Weakly communicating MDPs: }In this paper, we focused on the setting of finite horizon MDPs. Some previous approaches to exploration provide regret bounds for the more general setting of weakly communicating MDPs~\cite{jaksch2010near,bartlett2009regal}. However, we believe that our analysis can be extended to this more general case using existing techniques such as the ``doubling trick"~\cite{jaksch2010near}. 

\textbf{Computational and statistical efficiency: }The proposed algorithm is computationally tractable. In each episode, it performs an optimistic value iteration with computational cost of the same order as solving a known MDP. Besides, the obtained regret bounds guarantee with a high probability the statistical efficiency of the algorithm. 

\textbf{Theoretical Regret Bound} In current work, we show the
OppUCRL2 and UCRL2 can achieve same bound in periodic square wave variation factor case. However, in the simulation, the opportunistic version has significant better result. This implies the bound of OppUCRL2 is not tight, at least under some circumstances. In existing literature of finite horizon MDP, the analysis consider all state-action pairs together to get an upper bound of regret, which ignores difference of the strength of  the exploration.  So, in order to get a better bound for the opportunistic reinforcement learning algorithm, such as a better bound in opportunistic bandit setting \cite{Wu2018_AdaUCB} and \cite{Guo2019AdaLinUCBOL}, we may need to consider each state-action pair independently and we will consider this direction in the future work.

\end{document}